\documentclass{article}
\usepackage{amsmath, amssymb}
\usepackage{booktabs}   
\usepackage{multirow} 
\usepackage[preprint]{corl_2026}
\usepackage{graphicx}

\title{Invertible Neural Network Adapter for One-Step Flow Matching in Robot Manipulation}

\author{
  Yu~Zhang \And Kangyi ~Ji
 \And Yongxiang ~Zou \And Rongtao ~Xu \And Feng ~Zheng \And Long ~Cheng
}

\begin{document}
\maketitle


\begin{abstract}
This paper presents an invertible neural network adapter for general robotic manipulation, designed to generate precise high-dimensional actions conditioned on multimodal observations, including visual, linguistic, and proprioceptive inputs, through a one-step denoising process. Built upon a flow-matching formulation, the proposed adapter effectively constrains the action generation trajectory within an invertible latent space, thereby enabling efficient and high-quality dexterous action synthesis with only a single inference step. Compared with conventional iterative flow-matching policies, the proposed framework substantially reduces inference complexity while maintaining strong action prediction accuracy and stability. Extensive experiments are conducted across a diverse set of simulation benchmarks and real-world robotic platforms to evaluate the effectiveness of the proposed method. Across simulation benchmarks, the proposed adapter consistently demonstrates superior or near state-of-the-art performance on a wide range of manipulation tasks. Furthermore, real-world experiments reveal a significant improvement in inference efficiency for vision-language-action (VLA) models, reducing the average inference latency from 110 ms to 61 ms while maintaining strong task performance.
\end{abstract}

\keywords{Flow Matching, Robots Manipulation, Invertible Neural Network} 


\section{Introduction}

Generating reliable, precise, and dexterous actions in unstructured environments remains a fundamental challenge in robot learning\citep{wang2026dexjoco,sun2026maniparena,zhang2024navid,han2025multimodal}. Recent advances in diffusion-based policy learning \citep{chi2025diffusion} have significantly improved the modeling of high-dimensional and multimodal action distributions, leading to substantial progress in robotic manipulation\citep{hu2026anyslot,zhang2026a1,xu2025a0,ma2025phyblock,zhang2025robridge}. More recently, flow matching \citep{lipman2022flow} has emerged as an alternative generative modeling paradigm, demonstrating superior training efficiency and competitive performance in policy learning tasks \citep{chisari2024learning, black2024pi_0} compared with conventional diffusion-based approaches.
\begin{figure}[ht]
	\centering
	\includegraphics[width=\linewidth]{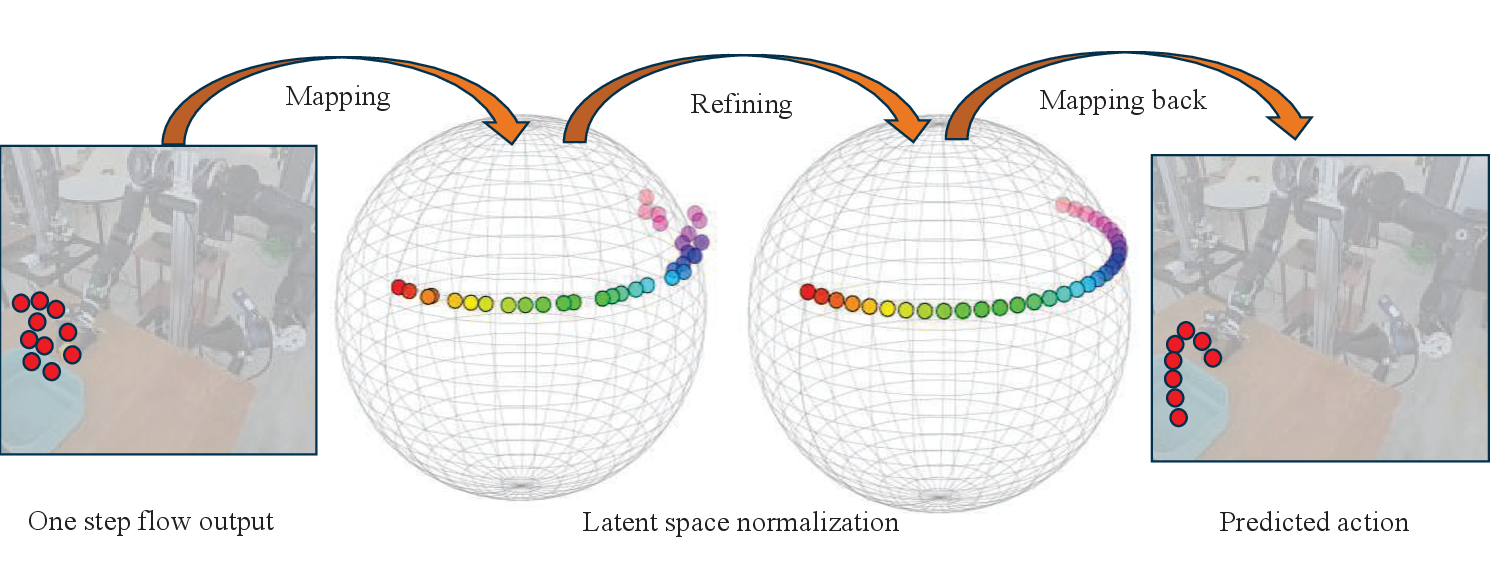}
	\caption{Overall framework of the proposed method}
	\label{fig:structure}
\end{figure}
Despite these advances, existing flow-matching-based policies \citep{chisari2024learning, black2024pi_0, braun2024riemannian, zhang2024affordance} still exhibit several limitations in efficiency, robustness, and generalization when deployed in complex real-world dexterous manipulation scenarios. In particular, current methods often struggle to accurately capture the fine-grained dynamics of multi-finger interactions, maintain temporal consistency over long-horizon action sequences, and generalize effectively to unseen environments and object configurations. Furthermore, most existing architectures lack a principled mechanism for integrating heterogeneous multimodal observations, including visual inputs, language instructions, and proprioceptive feedback, all of which are essential for robust real-world robotic manipulation.

Recent work in visuomotor imitation learning \citep{yan2025maniflow} extends the standard flow matching objective by introducing an additional consistency regularization term that enforces stable mappings from perturbed inputs to the target distribution, thereby enabling accurate action generation with substantially fewer inference steps. In contrast, \citep{geng2025mean} proposes to directly learn the mean velocity field for one-step generation, implicitly constraining velocity variation through the objective formulation. Although these approaches demonstrate promising empirical performance, they rely on introducing multiple auxiliary loss terms to approximate one-step generation. Consequently, while strong fitting performance can often be achieved on the training distribution, empirical observations suggest that the learned dynamics do not faithfully correspond to the true underlying mean velocity field during deployment, leading to degraded robustness and generalization in unseen scenarios.

To address these limitations, an invertible neural network (INN) adapter is proposed, in which the learned velocity field is constrained within an implicit invertible latent space. As illustrated in Fig.~\ref{fig:structure}, the proposed adapter is seamlessly integrated into the one-step flow matching framework. By the bijective property of invertible neural networks, action predictions generated by the original one-step flow matching model are refined, while information is preserved throughout the transformation process. In this manner, the irreversible loss of latent information is prevented, and the structural consistency of the underlying flow dynamics is maintained, thereby alleviating the approximation errors introduced by single-step inference. Consequently, the structural properties of the flow dynamics are preserved while efficient and stable one-step action generation is enabled. Furthermore, a more faithful alignment between the learned velocity field and the underlying action distribution is achieved, leading to improved robustness, enhanced temporal consistency, and stronger generalization across diverse robotic manipulation tasks.

Extensive experiments are conducted in both simulation and real-world environments to evaluate the effectiveness and generality of the proposed approach. In simulation, five manipulation tasks from the RoboTwin benchmark are evaluated using both RGB image observations and 3D point cloud observations. Furthermore, four tasks from the Libero benchmark are employed to assess large-scale vision-language-action learning, where a Qwen-VL-3-based vision-language model with a flow-matching action head is enhanced by the proposed adapter. Real-world evaluations are performed on three dexterous manipulation tasks using robotic arms.

The experimental results demonstrate that the proposed method consistently achieves state-of-the-art or near state-of-the-art performance across a wide range of manipulation tasks, observation modalities, and policy architectures. Notably, the proposed adapter not only improves policy accuracy and task success rates but also substantially reduces inference latency, making flow-matching policies more practical for real-time robotic manipulation.

\section{Method}
\subsection{Preliminaries: Flow Matching}
Following \citep{liu2023flow}, the forward process of the flow ODE is defined as a straight-line path between the data distribution and the noise distribution. Specifically, given a data sample $\mathbf{x}_1 \sim \mathcal{D}$, a noise sample $\mathbf{x}_0 \sim \mathcal{N}(\mathbf{0}, \mathbf{I})$, and a time step $t \sim \mathcal{U}[0,1]$, the interpolated state $\mathbf{x}_t$ is defined as
\begin{equation}
	\mathbf{x}_t = (1 - t)\mathbf{x}_0 + t\mathbf{x}_1.
\end{equation}
The corresponding velocity is defined as the displacement from the noise sample to the data sample:
\begin{equation}
	\mathbf{v}_t = \mathbf{x}_1 - \mathbf{x}_0.
\end{equation}
The flow model parameterized by $\theta$ is trained to predict the velocity given the noisy sample $\mathbf{x}_t$ at time $t$. The flow matching loss is formulated as
\begin{equation}
	\mathcal{L}_{\mathrm{FM}}(\theta) = \mathbb{E}_{\mathbf{x}_0 \sim \mathcal{N}(\mathbf{0}, \mathbf{I}),\, \mathbf{x}_1 \sim \mathcal{D},\, t \sim \mathcal{U}[0,1]}
	\left[ \left\| \mathbf{v}_{\theta}(\mathbf{x}_t, t) - (\mathbf{x}_1 - \mathbf{x}_0) \right\|^2 \right].
\end{equation}

\subsection{Preliminaries: Invertible Neural Network }

Following \citep{dinh2014nice}, invertibility is achieved by constructing an invertible neural network based on coupling layers. Specifically, the input vector $\mathbf{x} \in \mathbb{R}^D$ is partitioned into two disjoint subsets: 
\begin{equation} 
	\mathbf{x} = (\mathbf{x}_a, \mathbf{x}_b), 
	\end{equation}
 where $\mathbf{x}_a \in \mathbb{R}^{d}$ and $\mathbf{x}_b \in \mathbb{R}^{D-d}$. A coupling transformation is then defined such that one subset remains unchanged, while the other is transformed conditioned on the unchanged part: 
 \begin{align} 
 	\mathbf{y}_a &= \mathbf{x}_a, \\ \mathbf{y}_b &= \mathbf{x}_b + f(\mathbf{x}_a), \end{align} 
 	where $f(\cdot)$ denotes a nonlinear function parameterized by a neural network. This design ensures exact invertibility, as the inverse transformation can be computed in closed form: 
 	\begin{align} 
 		\mathbf{x}_a &= \mathbf{y}_a, \\ \mathbf{x}_b &= \mathbf{y}_b - f(\mathbf{y}_a). 
 	\end{align} Moreover, the Jacobian matrix of this transformation is triangular: \begin{equation} \frac{\partial \mathbf{y}}{\partial \mathbf{x}} = \begin{bmatrix} \mathbf{I} & \mathbf{0} \\ \frac{\partial f(\mathbf{x}_a)}{\partial \mathbf{x}_a} & \mathbf{I} \end{bmatrix}, 
 	\end{equation} whose determinant is trivially computed as: 
 	\begin{equation} 
 		\det \left( \frac{\partial \mathbf{y}}{\partial \mathbf{x}} \right) = 1. \end{equation} 
 		This property enables efficient likelihood computation and stable training. By stacking multiple such coupling layers with alternating partitions, the model can represent complex, highly nonlinear, yet exactly invertible transformations.

\subsection{Invertible Neural Network Adapter }

Since its proposal, flow matching has been designed primarily for one-step prediction tasks, with its core objective being to learn the straight-line path  between the target data distribution and a pre-specified noise distribution. In essence, the fundamental output of flow matching is a vector field that guides the transformation from noise to data along this intended linear trajectory. However, in practical applications, the actual vector field generated by standard flow matching often deviates from the ideal straight-line path due to the complexity of real-world data distributions, which are typically high-dimensional, non-linear, and exhibit complex structural characteristics. Consequently, the one-step generation paradigm adopted by basic flow matching often fails to achieve satisfactory performance.

To address these limitations, ManiFlow~\citep{yan2025maniflow} introduces a consistency loss designed to enforce a key property, namely that different points should converge to each other in a single step. Another representative approach is Mean Flow~\citep{geng2025mean}, which tackles the performance bottleneck of one-step generation from a complementary perspective by focusing on the dynamics of the vector field. Specifically, Mean Flow imposes constraints on the acceleration of the transformation process, thereby regularizing the temporal variation of the vector field.

In general, ManiFlow~\citep{yan2025maniflow} implicitly approximates the mean flow by minimizing the discrepancy between outputs corresponding to different initial points, whereas Mean Flow ~\citep{geng2025mean} explicitly learns the mean flow through a dedicated objective function. In contrast, the proposed method employs an invertible neural network adapter to map entire trajectories into a compact representation in the latent space, effectively collapsing them into a single point, from which the mean velocity can be directly inferred.

From a flow matching formulation, the velocity field $\mathbf{v}(\mathbf{x}_t, t \mid \mathbf{o})$ is predicted given a noisy sample $\mathbf{x}_t$ at time $t$, conditioned on the observation $\mathbf{o}$. The corresponding one-step denoised estimate $\mathbf{\hat{x}}$ is then given by
\begin{equation}
	\mathbf{\hat{x}} = \mathbf{x}_t + (1 - t)\,\mathbf{v}(\mathbf{x}_t, t \mid \mathbf{o}).
\end{equation}

Then, $\mathbf{\hat{x}}$ is fed into an invertible neural network. Its output is normalized to the spherical space under the assumption that a single-step action can be mapped to this spherical space. Subsequently, the normalized output is fed back to the invertible neural network, which performs an inverse transformation and directly outputs the final action for robot manipulation. This process can be expressed as:
\begin{equation}
	\begin{aligned}
		\mathbf{y} &= g(\hat{\mathbf{x}}), \\
		\tilde{\mathbf{y}} &= \frac{\mathbf{y}}{\lVert \mathbf{y} \rVert_2}, \\
		\mathbf{x}_{\mathrm{pre}} &= g^{-1}(\tilde{\mathbf{y}}),
	\end{aligned}
\end{equation}
where $g(\cdot)$ denotes the forward process of the invertible neural network and $ g^{-1}(\cdot)$ denotes the inverse process of the invertible neural network.

Which means that the output from the invertible neural network adapter is directly the robot action, compared to the vector field from the normal flow matching. 


 The adapter loss function is designed as:
\begin{equation}
	\mathcal{L}_{\mathrm{adapter}}(\theta) = \mathbb{E}_{\mathbf{x}_0 \sim \mathcal{N}(\mathbf{0}, \mathbf{I}),\, \mathbf{x}_1 \sim \mathcal{D},\, t \sim \mathcal{U}[0,1]}
	\left[
	\left\| \mathbf{x}_{\mathrm{pre}}(\mathbf{x}_t, t) - \mathbf{x}_1 \right\|^2
	+
	\left\| \tilde{\mathbf{y}}(\mathbf{x}_t, t) - g(\mathbf{x}_1) \right\|^2
	\right].
\end{equation}

Here, $\mathbf{x}_{\mathrm{pre}}(\mathbf{x}_t, t)$ denotes the reconstructed sample obtained via the inverse mapping, while $\tilde{\mathbf{y}}(\mathbf{x}_t, t)$ represents the latent representation in the latent space. By jointly optimizing reconstruction in both the data space and the latent space, the invertible neural network enforces consistency between the forward and inverse processes, thereby improving training stability and performance.

The final training objective is given by:
\begin{equation}
	\mathcal{L}(\theta) = \mathcal{L}_{\mathrm{adapter}}(\theta) + \alpha \, \mathcal{L}_{\mathrm{FM}}(\theta),
\end{equation}
where $\alpha > 0$ is a hyperparameter that controls the trade-off between the adapter objective and the flow matching objective.


\section{Experimental Results}
\label{sec:result}

\subsection{Simulation Experiments}
Five dexterous manipulation tasks from RoboTwin 1.0~\citep{Yu2019MetaWorldAB} are selected to evaluate the manipulation capabilities of the proposed adapter. For 2D image observations, comparisons are conducted against ManiFlow, Diffusion Policy, and Flow Matching Policy, all built upon a shared ResNet-18 encoder to ensure a fair comparison. For 3D point cloud observations, the proposed method is further compared with ManiFlow, 3D Diffusion Policy, and 3D Flow Matching. All baseline algorithms are implemented using the open-source codebase provided by~\citep{yan2025maniflow}, ensuring consistency in experimental settings and implementation details.

 For 2D image observations, the proposed method is trained with 300 epochs, while 600 epochs are used for training under 3D point cloud observations. Additionally, for 3D point cloud observations, two implementation strategies are adopted: first, noise is initialized from an all-zero distribution rather than sampled from Gaussian noise; second, timestep embeddings are omitted during training and inference. These modifications follow the empirical design choices for improving stability and performance in the 3D setting.

The compared algorithms are set consistent with those in the paper ManiFlow \cite{yan2025maniflow}. All data are trained using one 4090 GPU, and the quantitative results are presented in Tab. \ref{tab:2d_results} and Tab. \ref{tab:3d_results}.
\begin{table}[htbp!]
	\centering
	\scriptsize  
	\setlength{\tabcolsep}{4pt} 
	\caption{Performance comparison on RoboTwin 2D dexterous manipulation tasks}
	\label{tab:2d_results}
	\begin{tabular}{l c *{6}{c}} 
		\toprule
		Algorithm & Inference Step & Pick & Diverse & Dual & Empty & Shoe & Average \\
		\midrule
		2D Diffusion Policy & 10 & $17.0 \pm 0.8$ & $36.3 \pm 2.4$ & $41.3 \pm 3.7$ & $42.0 \pm 1.6$ & $7.3 \pm 2.9$ & $28.8 \pm 2.3$ \\
		2D Flow Matching  & 10 & $15.3 \pm 1.9$ & $32.0 \pm 4.5$ & $43.0 \pm 0.0$ & $38.0 \pm 5.4$ & $7.3 \pm 1.7$ & $27.1 \pm 2.7$ \\
		\multirow{2}{*}{2D ManiFlow Policy}
		& 1  & $29.3 \pm 8.2$ & $31.0 \pm 2.6$ & $31.6 \pm 0.8$ & $39.6 \pm 1.5$ & $13.6 \pm 10.8$ & $29.0 \pm 4.7$ \\
		& 10 & $37.3 \pm 4.8$ & $37.0 \pm 1.6$ & $47.3 \pm 2.1$ & $63.7 \pm 1.2$ & $\mathbf{45.3 \pm 3.7}$ & $46.1 \pm 2.7$ \\
		\midrule
		Proposed Method & 1 & $\mathbf{49.0 \pm 4.6}$ & $\mathbf{47.6 \pm 0.8}$ & $\mathbf{48 \pm 0.6}$ & $\mathbf{65.3 \pm 1.7}$ & $29.3 \pm 1.5$ & $47.6 \pm 1.8$ \\
		\bottomrule
	\end{tabular}
\end{table}

As shown in Tab.~\ref{tab:2d_results}, the proposed adapter consistently enables the flow matching policy to achieve state-of-the-art or near state-of-the-art performance across nearly all tasks on the RoboTwin benchmark, while requiring only a single inference step. Remarkably, despite reducing the denoising process from 10 inference steps to a single-step generation procedure, the proposed method still improves the average task success rate by $3.2\%$ over the baseline. This result suggests that the proposed adapter not only preserves the expressive capacity of flow matching policies under aggressive inference acceleration, but also enhances policy optimization by constraining velocity prediction in the latent space. Such improvements demonstrate that the proposed method effectively alleviates the performance degradation typically associated with one-step generation, achieving a favorable balance between computational efficiency and policy expressiveness. Furthermore, these results validate that introducing geometric constraints through the invertible adapter leads to more accurate action generation and stronger generalization across diverse dexterous manipulation tasks.
\begin{figure}[t]
	\centering
	\includegraphics[width=\linewidth]{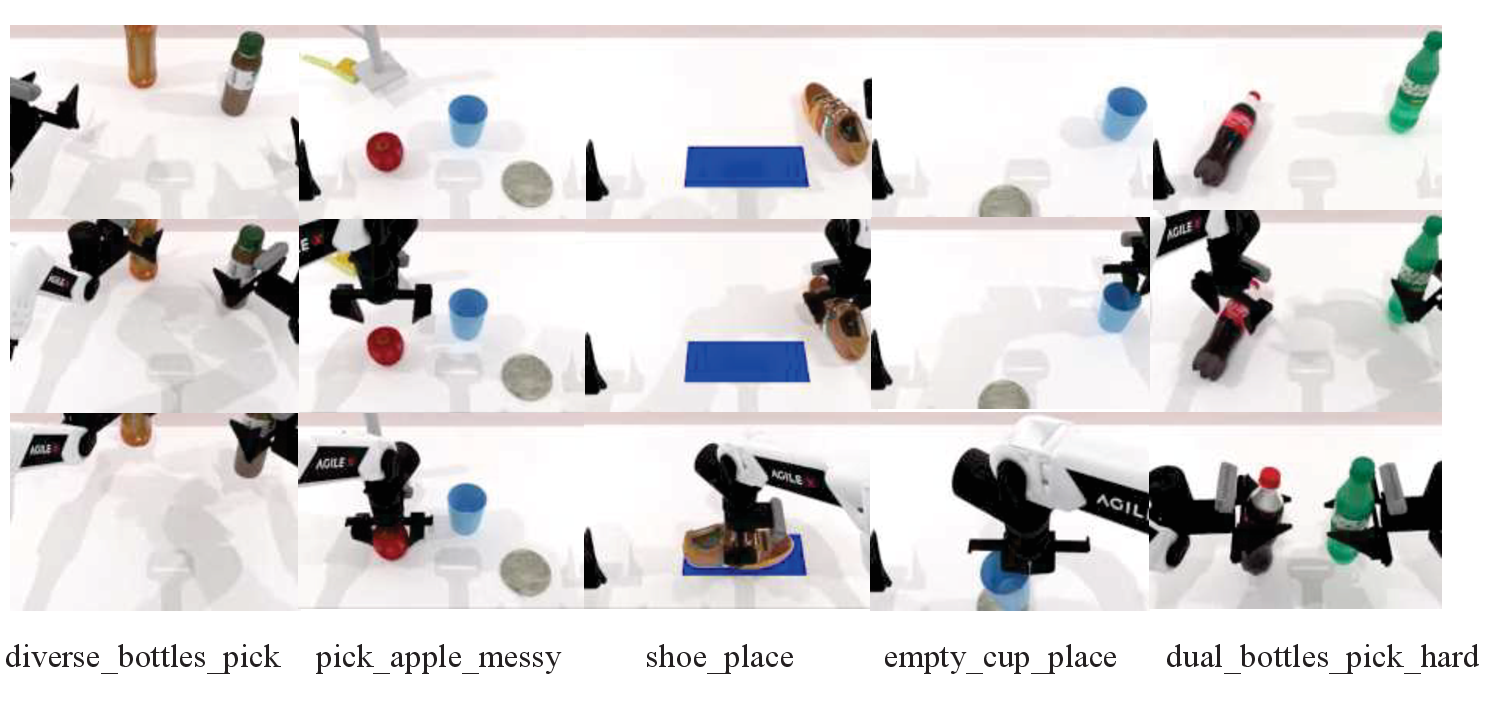}
	\caption{Chosen tasks on the RoboTwin benchmark}
	\label{fig:sim}
\end{figure}
\begin{table}[htbp!]
	\centering
	\scriptsize  
	\setlength{\tabcolsep}{4pt} 
	\caption{Performance comparison on RoboTwin 3D dexterous manipulation tasks}
	\label{tab:3d_results}
	\begin{tabular}{l c *{6}{c}} 
		\toprule
		Algorithm & Inference Step & Pick & Diverse & Dual & Empty & Shoe & Average \\
		\midrule
		3D Diffusion Policy & 10 & $9.3 \pm 3.7$ & $38.3 \pm 7.1$ & $46.3 \pm 2.5$ & $73.0 \pm 0.8$ & $46.5 \pm 2.5$ & $42.7 \pm 3.3$ \\
		3D Flow Matching  & 10 & $16.0 \pm 7.1$ & $56.3 \pm 6.6$ & $46.5 \pm 0.5$ & $82.3 \pm 1.7$ & $39.3 \pm 15.5$ & $48.1 \pm 6.3$ \\
		\multirow{2}{*}{3D ManiFlow Policy}
		& 1  & $42.7 \pm 1.9$ & $75.3 \pm 1.7$ & $53.7 \pm 0.5$ & $83.0 \pm 0.0$ & $63.7 \pm 2.6$ & $63.7 \pm 2.2$ \\
		& 10 & $42.0 \pm 0.8$ & $72.3 \pm 1.7$ & $54.0 \pm 2.2$ & $72.7 \pm 4.8$ & $68.3 \pm 2.9$ & $61.9 \pm 2.5$ \\
		\midrule
		Proposed Method & 1 & $\mathbf{54.6 \pm 0.8}$ & $65 \pm 0.6$ & $48.3 \pm 4.2$ & $\mathbf{88.3 \pm 1.5}$ & $\mathbf{68.6 \pm 0.2}$ & $64.9 \pm 1.4$ \\
		\bottomrule
	\end{tabular}
\end{table}
As shown in Tab.~\ref{tab:3d_results}, under 3D point cloud observations, the proposed method likewise achieves state-of-the-art or near state-of-the-art performance across nearly all RoboTwin tasks while maintaining a single inference step.  Demonstrating its robustness across observation modalities. In contrast to image observations, point cloud inputs explicitly encode geometric structure and positional information, which makes an all-zero initialization sufficient as the starting point for action generation. Based on this property, noise is initialized from zeros rather than sampled from a Gaussian distribution, simplifying the generation process without sacrificing performance. Moreover, because the zero initialization naturally corresponds to the starting state at $t=0$, explicit timestep embeddings become unnecessary, indicating that temporal conditioning can be omitted in this setting. This design further suggests that, for structured geometric observations, the flow generation process can be substantially simplified while preserving accurate action prediction. A quantitative comparison between the simplified architecture and the standard formulation is presented in Tab.~\ref{tab:compare}, where the simplified design demonstrates comparable or improved performance, further validating the effectiveness of these architectural modifications.
\begin{table}[htbp!]
	\centering
	\scriptsize  
	\setlength{\tabcolsep}{4pt} 
	\caption{Performance comparison on RoboTwin 3D dexterous manipulation tasks}
	\label{tab:compare}
	\begin{tabular}{l c *{6}{c}} 
		\toprule
		Algorithm  & Pick & Diverse & Dual & Empty & Shoe & Average \\
		\midrule
		Simplifying structure  & $\mathbf{54.6 \pm 0.8}$ & $65 \pm 0.6$ & $48.3 \pm 4.2$ & $\mathbf{88.3 \pm 1.5}$ & $\mathbf{68.6 \pm 0.2}$ & $64.9 \pm 1.4$ \\
		Normally structure  & $31.6 \pm 11.5$ & $\mathbf{71 \pm 0.6}$ & $\mathbf{53.3 \pm 1.5}$ & $78.6 \pm 1.5$ & $66 \pm 0.6$ & $60.1 \pm 3.1$ \\
		\bottomrule
	\end{tabular}
\end{table}

Furthermore, the proposed method significantly reduces the training cost. Specifically, the model converges within 300 training epochs, in contrast to the 2000 epochs required by Maniflow. Despite this substantial reduction in training time, the proposed approach not only maintains but further improves performance.  

In addition to the RoboTwin benchmark, the proposed method is further evaluated on the Libero benchmark \cite{liu2023libero}, which comprises a diverse suite of language-conditioned manipulation tasks and serves as a challenging testbed for assessing policy generalization across varying task semantics and environmental configurations. The baseline models are built upon QwenVL-3-4B \cite{Qwen3-VL} and are equipped with either a Pi-style flow matching action head or a Groot-style action head for action generation.

For the proposed approach, an invertible adapter module is inserted after the flow matching action head to further refine the generated action representations. To effectively incorporate high-level semantic information, the conditional tokens used by the adapter are extracted from the features of the last three transformer layers of QwenVL-3-4B. This design enables the adapter to leverage rich visual-language representations while preserving the efficiency of the underlying one-step flow matching framework, thereby enhancing action prediction accuracy and task execution performance.

All model parameters are trained using four NVIDIA A100 GPUs. Training is conducted for 8 epochs with a batch size of 32, resulting in a total training time of approximately 5--10 hours, depending on the task and dataset configuration. To ensure a fair comparison, model selection is performed based on evaluation performance, and the checkpoint achieving the highest validation score within the 5 training epochs is used for reporting the final results. For the baseline methods, the reported performance is directly adopted from \cite{community2026starvla} to maintain consistency with the original experimental protocol.

The quantitative results are presented in Tab.~\ref{tab:libero}. It can be observed that consistent performance improvements are obtained by incorporating the proposed adapter into the flow matching policy. These results indicate that more informative action representations can be learned through the proposed adapter, leading to enhanced task execution performance across a wide range of manipulation tasks. Notably, the performance gains are achieved without introducing additional flow matching inference steps, demonstrating that the proposed approach improves policy effectiveness while preserving inference efficiency.
\begin{table}[htbp!]
	\centering
	\scriptsize  
	\setlength{\tabcolsep}{4pt} 
	\caption{Performance comparison on Libero dexterous manipulation tasks}
	\label{tab:libero}
	\begin{tabular}{l c *{6}{c}} 
		\toprule
		Algorithm & Inference Step & Spatial & Object & Goal & Long  & Average \\
		\midrule
		QWen3+PI & 4 & $98.8$ & $\mathbf{99.6} $ & $95.8 $  & $88.4$ & $95.7$ \\
		\midrule
		QWen3+GROOT & 4 & $97.8$ & $98.8 $ & $97.4 $  & $92.0$ & $96.5$ \\
		\midrule
		QWen3+Flow matching+adapter & 1 & $\mathbf{99.0}$  & $\mathbf{99.6}$ & $\mathbf{97.6}$ & $\mathbf{93.6}$ & $\mathbf{97.4}$ \\
		\bottomrule
	\end{tabular}
\end{table}

Overall, the method achieves a favorable trade-off between computational efficiency and policy quality, reducing both training and inference overhead while delivering superior or comparable results.

\subsection{Real World Experiments}

To further validate the effectiveness of the proposed invertible adapter in real-world robotic manipulation, three representative tasks are evaluated: ``fruit picking and placement into a basket", ``block stacking", and ``dual-arm bottle grasping". The experiments are conducted using a combination of ``UR robotic" and the ``OpenArm platform", covering both single-arm and bimanual manipulation scenarios with varying levels of dexterity and coordination requirements.

\begin{figure}[t]
	\centering
	\includegraphics[width=\linewidth]{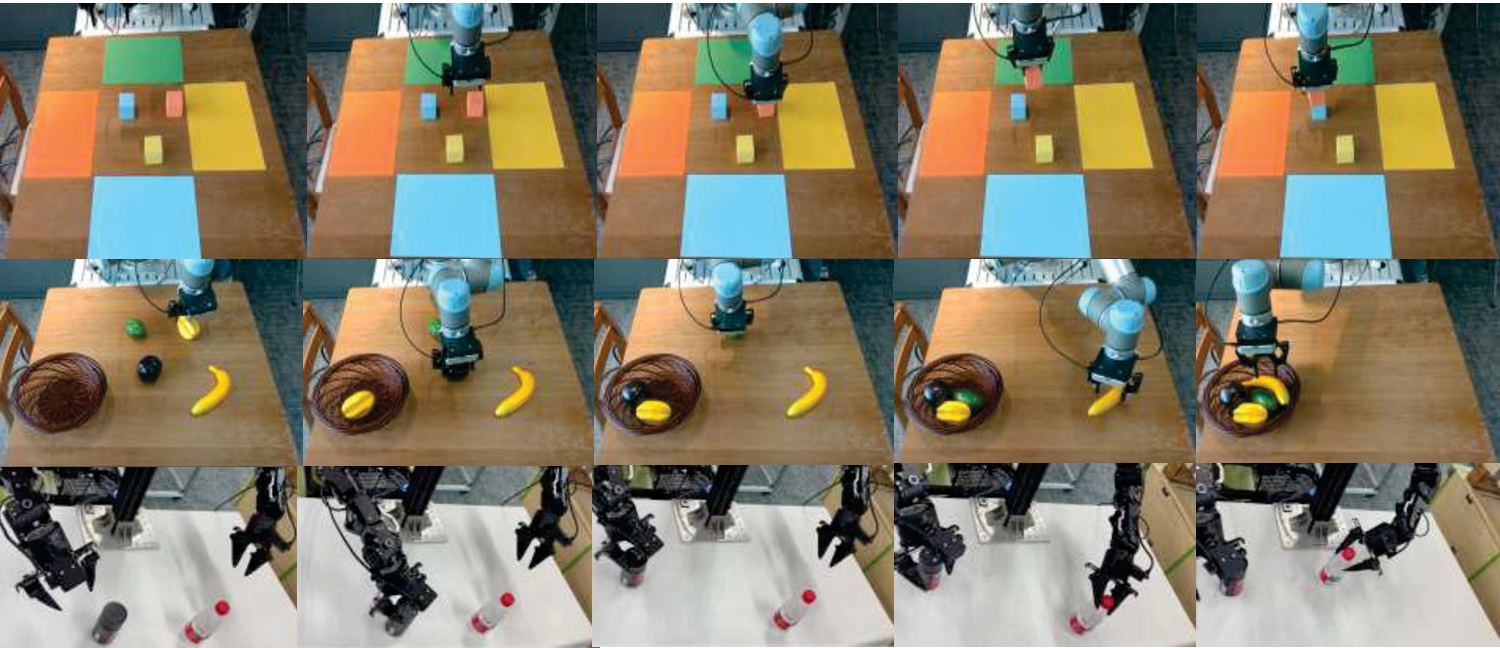}
	\caption{Real-world robotic manipulation tasks}
	\label{fig:real}
\end{figure}

As shown in Fig.~\ref{fig:real}, snapshots of the real-world experiments are presented. From top to bottom, the figure illustrates: block stacking using the UR platform, fruit picking and placement into a basket using the UR platform, and dual-arm bottle grasping using the OpenArm platform.

The original Pi 0.5 model \cite{black2025pi_05} is adopted as the baseline for comparison. To ensure a fair and controlled evaluation, both the baseline and the proposed ``Pi 0.5 + Adapter" model are trained using the same demonstration dataset, optimization strategy, training schedule, and hyperparameter settings. Consequently, any performance differences can be attributed solely to the introduction of the proposed adapter module.

Table~\ref{tab:realcompare} summarizes the success rates achieved on the three real-world manipulation tasks. Both methods achieve perfect performance on the fruit-picking and dual-arm bottle-grasping tasks, demonstrating that the proposed adapter preserves the strong manipulation capability of the original policy. On the more challenging block-stacking task, which requires higher precision in object alignment and placement, the proposed method improves the success rate from $80\%$ to $90\%$. This result indicates that the adapter can enhance the policy's action generation capability without degrading existing performance.
\begin{table}[htbp!]
	\centering
	\scriptsize  
	\setlength{\tabcolsep}{4pt} 
	\caption{Performance comparison on real manipulation tasks}
	\label{tab:realcompare}
	\begin{tabular}{l c *{3}{c}} 
		\toprule
		Algorithm  & pick fruits to basket & stack blocks & two arms grasp bottles  \\
		\midrule
		Pi0.5  & 10/10 & 8/10 &  10/10   \\
		Pi0.5+adapter  & 10/10 & 9/10 & 10/10  \\
		\bottomrule
	\end{tabular}
\end{table}
Moreover, the average inference time of the original Pi 0.5 model is approximately 110 ms per action prediction, requiring 10 denoising steps during inference. In contrast, the proposed method achieves action generation with only a single inference step, reducing the average inference time to approximately 61 ms. This corresponds to a reduction of nearly $45\%$ in inference latency, demonstrating the computational efficiency of the proposed adapter.

Overall, the experimental results demonstrate that the proposed adapter can be seamlessly integrated into the Pi 0.5 framework while consistently maintaining or improving task performance across diverse real-world manipulation scenarios. In particular, the performance improvement observed in the precision-sensitive block-stacking task suggests that the adapter enhances the expressiveness and adaptability of the learned action representation. At the same time, the substantial reduction in inference time highlights its ability to improve computational efficiency, making the approach more suitable for real-time robotic manipulation applications where both control accuracy and low-latency decision making are critical.

\section{Conclusion}
\label{sec:conclusion}
In this paper,  an invertible neural network adapter is presented for one-step flow matching in robotic manipulation. Unlike conventional flow-matching policies that directly learn action generation in the original action space, the proposed method leverages an invertible transformation to construct a task-adaptive latent action manifold. This design allows the flow model to perform generation in a more structured representation space while ensuring lossless reconstruction of the original actions through the inverse mapping. The proposed adapter can be seamlessly integrated into existing flow-matching architectures and vision-language-action models, improving both policy expressiveness and learning efficiency. Extensive experiments in simulation and real-world manipulation tasks demonstrate that the proposed approach consistently outperforms diffusion-based and existing flow-matching methods while maintaining low inference latency suitable for real-time robotic control.

\section{Limitation}
\label{sec:imitation}
While the invertible neural network adapter improves performance across both simulated and real-world manipulation tasks, several limitations remain. First, the method assumes that the action distribution can be mapped to a well-structured latent space via an invertible transformation. Although this holds for the manipulation tasks studied here, its validity for highly multimodal or discontinuous action distributions remains an open question. Second, our evaluation focuses on short-horizon tasks—grasping, object relocation, and dexterous interaction—and does not assess long-horizon tasks that demand complex sequential reasoning, hierarchical planning, or extensive memory. Third, although the adapter integrates with existing flow-matching and vision-language-action architectures, the invertible transformation introduces additional training-time computational and memory overhead. While inference remains efficient, further optimization will likely be needed to scale the method to larger foundation models. Addressing these limitations is essential for improving the scalability, generalization, and real-world applicability of one-step flow-matching policies.


\bibliography{example}  

\end{document}